%% file: main.tex
\documentclass[10pt,twocolumn,letterpaper]{article}

\usepackage{cvpr}              

\usepackage{graphicx}
\usepackage{amsmath}
\usepackage{amssymb}
\usepackage{booktabs}
\usepackage{multirow}
\usepackage{pifont}
\usepackage{nicefrac}
\usepackage{comment}
\usepackage{bbding}
\usepackage[pagebackref,breaklinks,colorlinks]{hyperref}

\mathchardef\mhyphen="2D

\usepackage{xcolor}

\usepackage[capitalize]{cleveref}
\crefname{section}{Sec.}{Secs.}
\Crefname{section}{Section}{Sections}
\Crefname{table}{Table}{Tables}
\crefname{table}{Tab.}{Tabs.}

\newcommand{\tablestyle}[2]{\setlength{\tabcolsep}{#1}\renewcommand{\arraystretch}{#2}\centering\footnotesize}
\newlength\savewidth\newcommand\shline{\noalign{\global\savewidth\arrayrulewidth
		\global\arrayrulewidth .8pt}\hline\noalign{\global\arrayrulewidth\savewidth}}

\def\cvprPaperID{840} 
\def\confName{CVPR}
\def\confYear{2022}

\begin{document}

\title{PYSKL: Towards Good Practices for Skeleton Action Recognition}
\author{Haodong Duan$^{1}$ \hspace{5mm} Jiaqi Wang$^{2}$ \hspace{5mm} Kai Chen$^{2}$ \hspace{5mm} Dahua Lin$^{1,2}$\\
$^{1}$The Chinese University of HongKong \hspace{8mm} $^{2}$Shanghai AI Laboratory}
\maketitle

\input{abstract.tex}
\input{intro.tex}
\input{gcn.tex}
\input{cnn.tex}

\input{conclusion.tex}

\bibliographystyle{ieee_fullname}
\bibliography{egbib}

\end{document}

%% file: abstract.tex
\begin{abstract}
    We present PYSKL: an open-source toolbox for skeleton-based action recognition based on PyTorch. 
    The toolbox supports a wide variety of skeleton action recognition algorithms, including approaches based on GCN and CNN.
    In contrast to existing open-source skeleton action recognition projects that include only one or two algorithms,
    PYSKL implements six different algorithms under a unified framework with both the latest and original good practices to ease the comparison of efficacy and efficiency.
    We also provide an original GCN-based skeleton action recognition model named ST-GCN++, 
    which achieves competitive recognition performance without any complicated attention schemes, serving as a strong baseline.
    Meanwhile, PYSKL supports the training and testing of nine skeleton-based action recognition benchmarks and achieves state-of-the-art recognition performance on eight of them.
    To facilitate future research on skeleton action recognition, we also provide a large number of trained models and detailed benchmark results to give some insights.
    PYSKL is released at https://github.com/kennymckormick/pyskl and is actively maintained.
    We will update this report when we add new features or benchmarks.
    The current version corresponds to PYSKL v0.2.
\end{abstract}

%% file: intro.tex
\section{Introduction}

Skeleton-based action recognition focuses on performing action recognition and video understanding based on human skeleton sequences.
Compared to other modalities (like RGB / Flow), skeleton data (2D / 3D human joint coordinates) are compact yet informative and robust to illumination changes or scene variations. 
Due to these good properties, skeleton action recognition has attracted increasing attention in recent years.
Various algorithms are developed to perform skeleton-based action recognition, which can be mainly categorized as GCN-based approaches and CNN-based approaches.

Since ST-GCN~\cite{yan2018spatial} first proposed to use Graph Convolutional Networks for skeleton processing,
GCN-based approaches soonly became the most popular paradigm in skeleton-based action recognition.
ST-GCN directly takes the sequence of joint coordinates as inputs and models skeleton data with a GCN backbone. 
The GCN backbone consists of alternating spatial graph convolutions and temporal convolutions for spatial and temporal modeling.
Following works~\cite{shi2019two,shi2020skeleton,liu2020disentangling,chen2021channel,zhang2020semantics,cheng2020skeleton,ye2020dynamic,li2019actional} inherited its basic design and made different improvements by: 
1) developing better graph topologies, either manual~\cite{liu2020disentangling} or learnable~\cite{shi2019two,ye2020dynamic,chen2021channel,zhang2020semantics}; 
2) training skeleton-based action recognition jointly with other auxiliary tasks~\cite{li2019actional}; 
3) adopting better data pre-processing, training, and testing strategies~\cite{shi2019two,shi2020skeleton,chen2021channel}. 
With these improvements, there comes great progress in recognition performance. 
For instance, on NTURGB+D-Xsub~\cite{shahroudy2016ntu} benchmark, the improvement of Top-1 Acc is over 10\%: from 81.5\% (ST-GCN~\cite{yan2018spatial}, 2018) to 92.4\% (CTR-GCN~\cite{chen2021channel}, 2021). 

Despite the considerable improvements, the settings of different GCN approaches do not align well.
For example, ST-GCN only reports the recognition performance with a single joint-stream, while most following works report the performance with an ensemble of joint / bone-stream (first proposed by 2s-AGCN~\cite{shi2019two}) or even four streams (first proposed by MS-AAGCN~\cite{shi2020skeleton}). 
Besides, the pre-processing techniques (skeleton alignments, temporal padding, denoising, \emph{etc.}) also differ a lot. 
However, existing open-source repositories~\cite{yan2018spatial,shi2020skeleton,liu2020disentangling,chen2021channel} only implement a single algorithm with their own practices.
To the best of our knowledge, none of them had ever compared these architectures under a unified setting.
Therefore, we developed PYSKL, which includes implementations of representative GCN approaches under a unified framework.
We trained and tested each algorithm with all the latest and original good practices on multiple benchmarks.
Surprisingly, we find that the recognition performance of different GCNs does not vary a lot: 
the extreme deviations of Top-1 Acc are less than 2\% on all four NTURGB+D benchmarks~\cite{shahroudy2016ntu,liu2020ntu}.
Especially, on NTURGB+D XView, the current state-of-the-art CTR-GCN~\cite{chen2021channel} only outperforms the original ST-GCN~\cite{yan2018spatial} by 0.5\%.

Given the pilot experiments, we find that good practices contribute more to achieving strong recognition performance rather than a complicated architectural design. 
This report presents all the good practices we adopted for training GCN-based models for skeleton-based action recognition. 
The practices include different aspects, including data pre-processing, spatial / temporal data augmentations, and hyper-parameter settings. 
Besides that, we also propose an original GCN algorithm named ST-GCN++.
With simple modifications on top of the ST-GCN, we achieve strong recognition performance comparable with the state-of-the-art without any complicated attention mechanism.
ST-GCN++ can serve as a strong baseline for future research on skeleton-based action recognition. 

Another paradigm for skeleton-based action recognition leverages Convolutional Neural Networks to process skeleton data. 
These approaches~\cite{choutas2018potion,yan2019pa3d,duan2021revisiting} represent human joints as Gaussian maps and 
aggregate them as pseudo images or video clips. The generated inputs are processed by 2D-CNN or 3D-CNN.
We implement a recent state-of-the-art 3D-CNN based approach PoseC3D~\cite{duan2021revisiting} in PYSKL.
PoseC3D can achieve strong recognition performance on skeleton-based action recognition benchmarks 
and has unique advantages (like robustness, scalability, interoperability) compared to GCNs. 
However, it is much heavier than most of the existing GCN approaches due to its 3D-CNN backbone.

To summarize, PYSKL implements six representative skeleton-based action recognition approaches and supports nine different benchmarks. 
It provides extensive benchmark results for five GCNs on skeleton-based action recognition, including four NTURGB+D benchmarks, four skeleton modalities,
and two annotation types (3D / 2D skeleton). 

PYSKL is released at https://github.com/kennymckormick/pyskl under the Apache-2.0 License. 
The repository contains all the source code, a large-scale model zoo, detailed instructions for installation, dataset preparation, and (distributed) training and testing.
PYSKL also provides tools for visualizing 2D / 3D skeletons and performing skeleton-based action recognition on custom datasets with no skeleton information available.

%% file: gcn.tex
\begin{table*}[t]
  \captionsetup{position=top}
  \caption{\textbf{Benchmarking GCN skeleton-based action recognition algorithms on the NTURGB+D benchmark. } Inputs are 3D skeletons with 25 joints. We set the input length to 100, input person number to 2, and apply all good practices introduced in Sec~\ref{sec-practice}. }
  \label{bm-ntu60}
  \vspace{-2mm}
  \resizebox{\linewidth}{!}{
  \tablestyle{8pt}{1.2}
      \begin{tabular}{c|cccc|cccc|c|c}
      \shline
       & \multicolumn{4}{c|}{NTURGB+D XSub} & \multicolumn{4}{c|}{NTURGB+D XView} & \multicolumn{2}{c}{Computational Efficiency} \\ \shline
      Model & Joint & Bone & 2s & 4s & Joint & Bone & 2s & 4s & GFLOPs & MParams \\ \shline
      ST-GCN~\cite{yan2018spatial} & 87.8 & 88.6 & 90.0 & 90.7 & 95.5 & 95.0 & 96.2 & 96.5 & 5.34 & 3.08 \\ \hline
      AAGCN~\cite{shi2020skeleton} & 89.0 & 89.2 & 90.8 & 91.5 & 95.7 & 95.2 & 96.4 & 96.7 & 6.07 & 3.77 \\ \hline
      MS-G3D~\cite{liu2020disentangling} & \textbf{89.6} & 89.3 & 91.0 & 91.7 & \textbf{95.9} & 95.0 & 96.4 & 96.9 & 10.27 & 3.17 \\ \hline
      CTR-GCN~\cite{chen2021channel} & \textbf{89.6} & 90.0 & \textbf{91.5} & \textbf{92.1} & 95.6 & 95.4 & 96.6 & \textbf{97.0} & 2.82 & 1.43 \\ \hline
      ST-GCN++ & 89.3 & \textbf{90.1} & 91.4 & \textbf{92.1} & 95.6 & \textbf{95.5} & \textbf{96.7} & \textbf{97.0} & \textbf{2.80} & \textbf{1.39} \\ \shline
      \end{tabular}}
      \vspace{-1mm}
\end{table*}

\begin{table*}[t]
  \captionsetup{position=top}
  \caption{\textbf{Benchmarking GCN skeleton-based action recognition algorithms on the NTURGB+D 120 benchmark. } Inputs are 3D skeletons with 25 joints. We set the input length to 100, input person number to 2, and apply all good practices introduced in Sec~\ref{sec-practice}. }
  \label{bm-ntu120}
  \vspace{-2mm}
  \resizebox{\linewidth}{!}{
  \tablestyle{8pt}{1.2}
      \begin{tabular}{c|cccc|cccc|c|c}
      \shline
       & \multicolumn{4}{c|}{NTURGB+D 120 XSub} & \multicolumn{4}{c|}{NTURGB+D 120 XSet} & \multicolumn{2}{c}{Computational Efficiency} \\ \shline
       Model & Joint & Bone & 2s & 4s & Joint & Bone & 2s & 4s & GFLOPs & MParams \\ \shline
       ST-GCN~\cite{yan2018spatial} & 82.1 & 83.7 & 85.6 & 86.2 & 84.5 & 85.8 & 87.5 & 88.4 & 5.34 & 3.08 \\ \hline
       AAGCN~\cite{shi2020skeleton} & 82.8 & 84.7 & 86.3 & 86.9 & 84.8 & 86.2 & 88.1 & 88.8 & 6.07 & 3.77 \\ \hline
       MS-G3D~\cite{liu2020disentangling} & \textbf{84.0} & 85.3 & 86.9 & 87.8 & \textbf{86.0} & 87.3 & 88.9 & 89.6 & 10.27 & 3.17 \\ \hline
       CTR-GCN~\cite{chen2021channel} & \textbf{84.0} & \textbf{85.9} & \textbf{87.5} & \textbf{88.1} & 85.9 & 87.4 & \textbf{89.2} & \textbf{89.9} & 2.82 & 1.43 \\ \hline
       ST-GCN++ & 83.2 & 85.6 & 87.0 & 87.5 & 85.6 & \textbf{87.5} & 89.1 & 89.8 & \textbf{2.80} & \textbf{1.39} \\ \shline
      \end{tabular}}
      \vspace{-1mm}
\end{table*}

\section{GCN-based approaches}

\subsection{Good Practices for GCN-based approaches}
\label{sec-practice}
\subsubsection{Data Pre-processing}
Skeleton sequences of different videos may have different temporal lengths, different numbers of persons, and may be captured by sensors from different views or with different setups.
For 3D skeletons~\cite{shahroudy2016ntu,liu2020ntu} captured with Kinect sensors~\cite{zhang2012microsoft}, various pre-processing approaches are adopted.
ST-GCN uses no extra pre-processing and pads all sequences to a maximum length with zero padding.
2s-AGCN, however, performs pre-normalization by 
1) aligning the center point of the person in the $1_{st}$ frame with the origin of the 3D-Cartesian coordinate system;
2) rotating all skeletons so that the spine of the person in the first frame is parallel with the $z$-axis in the 3D-Cartesian coordinate system.
Besides, 2s-AGCN pads skeleton sequences to a maximum length with loop padding.
CTR-GCN follows the spatial pre-processing used by 2s-AGCN. 
However, it keeps the original length of each skeleton sequence and uses different criteria for manual denoising.
PYSKL follows the pre-processing approach of CTR-GCN.
For 2D skeletons predicted by pose estimators~\cite{sun2019deep,8765346}, we pre-normalize them into a fixed range (like [0, 1] or [-1, 1]) following~\cite{yan2018spatial}. 
We also perform simple pose-based tracking to form 1 or 2 skeleton sequences for NTURGB+D data.

\subsubsection{Temporal Augmentations}
Most GCN works do not use any temporal augmentations. 
Among representative GCN approaches, CTR-GCN adopts random cropping as temporal augmentations.
It crops a substring from the entire skeleton sequence (substring length ratio may vary from 50\% to 100\%) 
and resize the substring to a fixed length of 64 with bilinear interpolation.
Inspired by \cite{duan2021revisiting}, we use Uniform Sampling as the temporal augmentation strategy.
To generate a skeleton sequence of length M (M=100 in PYSKL), we divide the original sequence uniformly into M splits with equal lengths and randomly sample one frame per split. 
The sampled skeletons will be joined again and form the input sequence. 
With Uniform Sampling, we can generate numerous data samples with similar distribution to the source data (no interpolation used). 

\subsubsection{Hyper Parameter Setting}
The hyper parameter settings differ a lot in previous works for skeleton-based action recognition using GCN. 
In PYSKL, we use the same hyper parameter setting to train all GCN models.
We set the initial learning rate to 0.1, batch size to 128, and train each model for 80 epochs with the CosineAnnealing LR scheduler.
For the optimizer, we set the momentum to 0.9, weight decay to $5\times 10^{-4}$, and use the Nesterov momentum.
We find that for most GCN networks, the new hyper parameter setting leads to better recognition performance than previous settings that use the MultiStep LR scheduler.

\subsection{The Design of ST-GCN++}
We also propose an original GCN model named ST-GCN++.
With only simple modifications to the original ST-GCN,
ST-GCN++ achieves strong recognition performance comparable with the state-of-the-art approach with a complicated attention mechanism. 
Meanwhile, the computational overhead is also reduced.
ST-GCN++ modifies the design of the interleaving spatial modules (spatial graph convolutions) and temporal modules (temporal 1D convolutions). 

\subsubsection{Spatial Module Design}
In ST-GCN, pre-defined sparse coefficient matrices are used for fusing features of different joints belonging to the same person, 
while the coefficient matrices are derived from a pre-defined joint topology. 
Meanwhile, ST-GCN also re-weights each element in coefficient matrices with a set of learnable weights.
However, in ST-GCN++, we only use the pre-defined joint topology to initialize the coefficient matrices.
We update the coefficient matrices iteratively with gradient descent during training without any sparse constraints. 
Besides, we also add a residual link in the spatial module, which further improves the spatial modeling capability.

\subsubsection{Temporal Module Design}
A vanilla ST-GCN uses a single 1D convolution on the temporal dimension with kernel size 9 for temporal modeling.
The large kernel covers a wide temporal receptive field. 
However, this design lacks flexibility and results in redundant computations and parameters. 
Inspired by \cite{liu2020disentangling,chen2021channel}, we use a multi-branch temporal ConvNet (TCN) to replace the single branch design.
The adopted multi-branch TCN consists of six branches: a `1x1' Conv branch, a Max-Pooling branch, and four temporal 1D Conv branches with kernel size 3 and dilations from 1 to 4.
It first transforms features with `1x1' Conv and divides them into six groups with equal channel width.
Then, each feature group is processed with a single branch.
The six outputs are concatenated together and processed by another `1x1' Conv to form the output of the multi-branch TCN.
The new TCN design not only improves the temporal modeling capabililty, but also saves the computational cost and parameters, due to the reduced channel width for every single branch.

\begin{table}[t]
  \captionsetup{position=top}
  \caption{\textbf{Benchmarking spatial augmentations using ST-GCN++ on two NTURGB+D 120 benchmarks. } Random rotation works for 3D skeletons, while random scaling works for both 2D and 3D skeletons. }
  \label{bm-aug}
  \vspace{-2mm}
  \resizebox{\linewidth}{!}{
  \tablestyle{4pt}{1.2}
      \begin{tabular}{c|c|ccc|ccc}
      \shline
        \multicolumn{2}{c|}{}  & \multicolumn{3}{c|}{NTURGB+D 120 XSub} & \multicolumn{3}{c}{NTURGB+D 120 XSet} \\ \shline
        Spatial Augs & Anno & Joint & Bone & 2s & Joint & Bone & 2s \\ \shline 
        None & 3D & 83.2 & 85.6 & 87.0 & 85.6 & 87.5 & 89.1 \\ \hline
        Rot & 3D & 83.8 & \textbf{86.0} & 87.7 & \textbf{86.8} & \textbf{88.0} & \textbf{89.9} \\ \hline
        Scale & 3D & 84.0 & 85.9 & 87.7 & 86.3 & 87.3 & 89.2 \\ \hline
        Rot + Scale & 3D & \textbf{84.7} & \textbf{86.0} & \textbf{87.9} & 86.6 & \textbf{88.0} & 89.8 \\ \shline
        None & 2D & 84.4 & 84.8 & 86.4 & 88.1 & 88.5 & 90.0 \\ \hline
        Scale & 2D & \textbf{85.1} & \textbf{85.7} & \textbf{87.1} & \textbf{88.7} & \textbf{90.0} & \textbf{90.9} \\ \shline
      \end{tabular}}
      \vspace{-2mm}
\end{table}

\begin{table}[t]
  \captionsetup{position=top}
  \caption{\textbf{ST-GCN++ trained with good practices and spatial augmenatations surpasses CTR-GCN (official performance) on 3 of 4 NTURGB+D benchmarks. }}
  \label{bm-cmp}
  \vspace{-2mm}
  \resizebox{\linewidth}{!}{
  \tablestyle{12pt}{1.2}
      \begin{tabular}{c|cc|cc}
      \shline
       & \multicolumn{2}{c|}{NTURGB+D} & \multicolumn{2}{c}{NTURGB+D 120} \\ \shline
      Model & XSub & XView & XSub & XSet \\ \shline
      CTR-GCN & 92.4 & 96.8 & \textbf{88.9} & 90.6 \\ \hline
      ST-GCN++ & \textbf{92.6} & \textbf{97.4} & 88.6 & \textbf{90.8} \\ \shline
      \end{tabular}}
      \vspace{-2mm}
\end{table}

\begin{table*}[th]
  \captionsetup{position=top}
  \caption{\textbf{The best performance achieved by PYSKL on nine benchmarks. } On 8 of 9 benchmarks for skeleton action recognition, PYSKL achieves the best recognition accuracy. 
  We report the Top-1 accuracy for all benchmarks except FineGYM (for which we report the mean class accuracy). For Diving48, we use the V2 annotations. }
  \label{bm-all}
  \vspace{-2mm}
  \resizebox{\linewidth}{!}{
  \tablestyle{4pt}{1.2}
      \begin{tabular}{c|cc|cc|c|c|c|c|c}
      \shline
       & \multicolumn{2}{c|}{NTURGB+D (3D)} & \multicolumn{2}{c|}{NTURGB+D 120 (3D)} & \multicolumn{5}{c}{Datasets with 2D skeleton annotations} \\ \shline
       & XSub & XView & XSub & XSet & Kinetics-400 & UCF101 & HMDB51 & FineGYM & Diving48 \\ \shline
      Previous SOTA & 92.4~\cite{chen2021channel} & 96.8~\cite{chen2021channel} & \textbf{88.9}~\cite{chen2021channel} & 90.6~\cite{chen2021channel} & 38.6~\cite{obinata2021temporal} & 69.1~\cite{yan2019pa3d} & 53.5~\cite{yan2019pa3d} & N.A. & N.A. \\ \hline
      PYSKL & \textbf{92.6} & \textbf{97.4} & 88.6 & \textbf{90.8} & \textbf{49.1} & \textbf{86.9} & \textbf{69.4} & \textbf{94.1} & \textbf{54.5} \\ \shline
  \end{tabular}}
  \vspace{-2mm}
\end{table*}

\subsection{Benchmarking GCN Algorithms}
In PYSKL, we benchmark four representative GCN approaches: ST-GCN~\cite{yan2018spatial}, AAGCN~\cite{shi2020skeleton}, MS-G3D~\cite{liu2020disentangling}, CTR-GCN~\cite{chen2021channel}, as well as the original ST-GCN++ on four NTURGB+D benchmarks~\cite{shahroudy2016ntu,liu2020ntu}. 
For skeleton annotations, we consider 3D skeletons generated with CTR-GCN pre-processing and 2D skeletons estimated by HRNet~\cite{sun2019deep}. 
We report the Top-1 Accuracy of joint-stream, bone-stream, two-stream fusion (joint + bone), and four-stream fusion (joint + bone + joint motion + bone motion), respectively.
This report lists the benchmark results when using 3D skeletons in Table~\ref{bm-ntu60},~\ref{bm-ntu120}.
The results for 2D skeletons can be found in the repository. 

Unlike the performance originally reported, we find that the accuracy gaps between different GCN approaches are much smaller.
On all NTURGB+D benchmarks, the extreme deviation of Top-1 Accuracy is less than 2\%. 
For all algorithms except CTR-GCN, the reproduced results are better than reported due to the adopted good practices\footnote{For CTR-GCN, the performance dropped a little, since spatial augmentations (like random rotation) are used in the original paper, but not used in this benchmark. }.
Moreover, our ST-GCN++ achieves strong recognition performance competitive with state-of-the-art GCN approaches, with a much simpler design, fewer parameters, and fewer FLOPs. 

\subsection{Spatial Augmentations}
We also adopt spatial augmentations in skeleton action recognition. 
We implement three augmentations in PYSKL: 

1) Random Rotating: Rotate all skeletons (2D / 3D) with the same random angle $\theta$ ($\theta = (\theta_x, \theta_y) \in \mathbf{R}^2$ or $\theta = (\theta_x, \theta_y, \theta_z) \in \mathbf{R}^3$),  
each element in $\theta$ is sampled from a uniform distribution [-0.3, 0.3].

2) Random Scaling: Scale all joint coordinates in a skeleton sequence with the same scale factor $r$ ($r = (r_x, r_y) \in \mathbf{R}^2$ or $r = (r_x, r_y, r_z) \in \mathbf{R}^3$), 
each element in $r$ is sampled from a uniform distribution [-0.1, 0.1] or [-0.2, 0.2] (for 3D / 2D skeletons).

3) Random Gaussian Noise: Randomly add a small Gaussian noise for each joint. The noise can be frame-specific or frame-agnostic.

Extensive experiments are conducted to validate the efficacy of three spatial augmentations (results in Table~\ref{bm-aug}). 
We find that among the three augmentations, random rotating works for 3D skeletons; random scaling works for both 2D and 3D skeletons; while random Gaussian noise does not work for any kinds of skeletons.
We train ST-GCN++ with 3D skeletons for 120 epochs with random rotating and random scaling.
Table~\ref{bm-cmp} shows that on 3 of 4 NTURGB+D benchmarks, ST-GCN++ surpasses the current state-of-the-art CTR-GCN.

%% file: cnn.tex
\section{CNN-based approaches}

PYSKL also implements the 3D-CNN based approach PoseC3D~\cite{duan2021revisiting}.
PoseC3D takes 2D human skeletons as inputs.
It first generates Gaussian maps given the 2D joint coordinates and then organizes them as a 3D heatmap volume. 
PoseC3D can use an arbitrary 3D-CNN for 3D heatmap volume processing. 
In PYSKL, we support three backbones: C3D~\cite{tran2015learning}, SlowOnly~\cite{feichtenhofer2019slowfast}, X3D~\cite{feichtenhofer2020x3d},
and release PoseC3D trained on seven different datasets: NTURGB+D~\cite{shahroudy2016ntu}, NTURGB+D 120~\cite{liu2020ntu}, Kinetics-400~\cite{carreira2017quo}, UCF101~\cite{soomro2012ucf101}, HMDB51~\cite{kuehne2011hmdb}, FineGYM~\cite{shao2020finegym}, and Diving48~\cite{li2018resound}.
PoseC3D has good spatio-temporal modeling capability and achieves state-of-the-art recognition performance on 6 of 9 benchmarks. 
However, using 3D-CNN for skeleton processing consumes more computations and is much slower than representative GCN approaches.

%% file: conclusion.tex
\section{Conclusion}
We have publicly released PYSKL, which has extensive benchmarks for skeleton-based action recognition. 
PYSKL has implemented six representative algorithms under a unified framework, trained them on nine different skeleton-based action recognition benchmarks, and achieved state-of-the-art recognition performance on eight benchmarks (Table~\ref{bm-all}). 
It has offered good practices for training skeleton-based action recognition models and provided extensive benchmarks.
Besides, it also introduced a simple and strong baseline named ST-GCN++, which surpasses previous state-of-the-art on the NTURGB+D benchmarks. 
We hope this repository, along with all the released training configurations and model weights will facilitate future research in this area. 